\documentclass[]{ceurart}

\usepackage{listings}
\usepackage{graphicx} 
\usepackage{amsmath}
\usepackage{amssymb}
\usepackage{algorithm}
\usepackage{algpseudocode} 
\usepackage{multicol} 
\usepackage{enumitem}

\usepackage{alltt}
\usepackage{tabularx}

\begin{document}

% \copyrightyear{2025}
% \copyrightclause{Copyright for this paper by its authors.
%   Use permitted under Creative Commons License Attribution 4.0
%   International (CC BY 4.0).}

\conference{Accepted to LLAIS 2025: Workshop on LLM-Based Agents for Intelligent Systems, at ECAI 2025}

% \title{Identity Retrieval-Augmented Generation \\ for Generative Agent Identity Coherence}
% \title{Identity Retrieval-Augmented Generation \\ 
% for Generative Human-AI Agents}

\title{ID-RAG: Identity Retrieval-Augmented Generation for Long-Horizon Persona Coherence in Generative Agents}

% \tnotetext[1]{You can use this document as the template for preparing your
%   publication. We recommend using the latest version of the ceurart style.}

%%%% AUTHOR BLOCK %%%%
\author[1,2]{Daniel Platnick}[%
%orcid=0000-0002-000-0001,
email=daniel.platnick@flybits.com,
]
\cormark[1]
% \address[1]{Flybits Labs, Creative Ai Hub, Toronto, Canada}
% \address[2]{Toronto Metropolitan University, Toronto, Canada}
\address[1]{Flybits Labs, Creative Ai Hub}
\address[2]{Toronto Metropolitan University}

\author[1,2]{Mohamed E. Bengueddache}[%
%orcid=0000-0001-000-0000,
email=mohamed.bengueddache@flybits.com,
]

% \author[1,2]{Marjan Alirezaie}
% %\fnmark[1]

% \author[4]{\newline Dava J.
% Newman}

% \author[3,5]{Alex ``Sandy'' Pentland}

% \author[1,2,3]{Hossein Rahnama}
% % \address[3]{MIT Media Lab, Cambridge, USA}
% % \address[4]{Stanford University, California, USA}
% \address[3]{MIT Media Lab}
% \address[4]{Massachusetts Institute of Technology}
% \address[5]{Stanford University}
% % Footnotes
% \cortext[1]{Corresponding author.}

\author[1,2]{Marjan Alirezaie}
% \author[1,2]{Marjan Alirezaie\texorpdfstring{\\}{ }}


\author[3]{\texorpdfstring{\\}{ } Dava J.
Newman}

\author[3,4]{Alex ``Sandy'' Pentland}

\author[1,2,3]{Hossein Rahnama}
% \address[3]{MIT Media Lab, Cambridge, USA}
% \address[4]{Stanford University, California, USA}
\address[3]{Massachusetts Institute of Technology}
\address[4]{Stanford University}
% \address[5]{MIT Media Lab}

% Footnotes
\cortext[1]{Corresponding author.}

%%%% END AUTHOR BLOCK %%%%

%\fntext[1]{These authors contributed equally.}

\begin{abstract}
% Generative agents powered by language models are increasingly deployed for long-horizon tasks.
% However, as long-term memory context grows over time, they struggle to maintain a coherent persona.
% This deficiency leads to critical failures, including identity drift, ignoring established beliefs, and the propagation of hallucinations in multi-agent systems.
% To mitigate these challenges, this paper introduces Identity Retrieval-Augmented Generation (ID-RAG), a novel mechanism designed to ground agent self-perception in a persistent and coherent sense of self.
% ID-RAG grounds agent behavior in a dynamic, structured identity model: a knowledge graph of core beliefs, traits, and values. 
% During the agent's decision loop, this model is queried to retrieve relevant identity context, which directly informs action selection. 
Generative agents powered by language models are increasingly deployed for long-horizon tasks.
However, as long-term memory context grows over time, they struggle to maintain coherence.
This deficiency leads to critical failures, including identity drift, ignoring established beliefs, and the propagation of hallucinations in multi-agent systems.
To mitigate these challenges, this paper introduces \textbf{Identity Retrieval-Augmented Generation (ID-RAG)}, a novel mechanism designed to ground an agent's persona and persistent preferences in a dynamic, structured identity model: a knowledge graph of core beliefs, traits, and values.
During the agent's decision loop, this model is queried to retrieve relevant identity context, which directly informs action selection.
We demonstrate this approach by introducing and implementing a new class of ID-RAG enabled agents called Human-AI Agents (HAis), where the identity model is inspired by the Chronicle structure used in Perspective-Aware AI—a dynamic knowledge graph learned from a real-world entity's digital footprint. 
% Social simulation experiments on HAis implemented in the Concordia Generative Agent-Based Modeling framework suggest ID-RAG significantly improves persona persistence and coherence, leading to more productive agent-to-agent interactions and, as a result, enhanced simulation fidelity. 
In social simulations of a mayoral election, HAis using ID-RAG outperformed baseline agents in long-horizon persona coherence—achieving higher identity recall across all tested models by the fourth timestep—and reduced simulation convergence time by 19\% (GPT-4o) and 58\% (GPT-4o mini).
By treating identity as an explicit, retrievable knowledge structure, ID-RAG offers a foundational approach for developing more temporally coherent, interpretable, and aligned generative agents. 
Our code is open-source and available at: \url{https://github.com/flybits/humanai-agents}.
\end{abstract}


\maketitle
% Page numbers only (centered), no headers — no external packages
\makeatletter
\def\ps@plain{%
  \let\@oddhead\@empty
  \let\@evenhead\@empty
  \def\@oddfoot{\hfil\thepage\hfil}%
  \let\@evenfoot\@oddfoot
}
\makeatother
\pagenumbering{arabic}
\thispagestyle{plain}  % number the title page
\pagestyle{plain}      % use our plain style everywhere

\section{Introduction}
\label{sec:introduction}

Driven by the advanced reasoning and natural language capabilities of modern language models, generative agents are rapidly advancing the frontier of artificial intelligence \cite{luo2025large_language_model_agent_survey,plaat2025agentic_llm_survey}.
Their applications range from creating believable social simulacra for computational social science \cite{Park-simulacra-2023,Vezhnevets-concordia-2023} to executing long-horizon tasks as autonomous assistants \cite{erdogan2025planandactimprovingplanningagents}. 
This progress is rooted in the ability of generative agents to perceive, reason, and act within complex, dynamic environments, often yielding
% remarkably
human-like behaviors \cite{Gao-llmsurvey-2023,shao2023character_llm,liang2025mars_memory_agents}.

However, maintaining persona coherence remains a fundamental challenge, threatening the long-term viability of these agents.
During extended interactions, agents often suffer identity drift, causing their foundational persona to degrade. 
This drift leads to critical failures: agents develop behavioral contradictions, become more susceptible to influence, and experience self-perceptive hallucinations which can be propagated to other agents through agent-to-agent interactions \cite{li2025personalization_survey,zheng2025lifelong_learning_roadmap,chen2024relyllmagentsdraft,Lu-llmhallucination-2024}.

This paper introduces Identity Retrieval-Augmented Generation (ID-RAG), a mechanism designed to equip generative agents with a stable persona and persistent preferences. 
Instead of representing an agent's persona through an implicit, transient state within a long-term memory module, ID-RAG employs an explicit, structured identity model \cite{qiao2024self_evolving_agents,qiao2025agentic_self_awareness,wang2024craftingpersonalizedagentsretrievalaugmented}. 
Under this paradigm, an agent's persona is grounded in a dynamic identity knowledge graph of beliefs, traits, and values. 
To ensure decisions consistently align with this core self-representation, actions are informed by targeted identity retrieval from the graph and can be validated by an optional gating mechanism.

To demonstrate and evaluate ID-RAG in a practical setting, we introduce the Human-AI Agent (HAi) architecture.
In our implementation, HAis operationalize ID-RAG by retrieving from a Chronicle—an identity knowledge graph inspired by Perspective-Aware AI (PAi) \cite{PAi-2021,Alirezaie-pai1-2024,alirezaie-pai2-2024,Platnick2025PAiXR,platnick-pai3-2024}—as their identity model.
For this study, our implementation uses static, handcrafted Chronicles and is deployed in the Concordia Generative Agent-Based Modeling (GABM) framework \cite{Vezhnevets-concordia-2023}.
This setup allows us to directly test our central hypothesis: grounding agents in an explicit identity model with ID-RAG improves long-horizon persona coherence compared to the baseline architecture from \cite{Park-simulacra-2023}.

The main contributions of this paper are as follows:

\begin{enumerate}
    \item We formalize ID-RAG, a mechanism designed to improve long-horizon persona coherence and alignment in generative agents. 
    It achieves this by equipping agents with a distinct identity model—a dynamic knowledge graph—enabling them to retrieve identity context to inform and optionally validate their actions.

    \item We introduce HAis, a new class of generative agents whose defining feature is using the ID-RAG mechanism to ground self-perception in an identity model that adopts the Chronicle structure from PAi.
    HAis align agent personas with real-world entities by performing ID-RAG on user Chronicles derived from entity-specific digital footprint data.
    We provide a baseline HAi implementation in Concordia, demonstrating its use of ID-RAG on manually constructed Chronicles to align agent personas.

    \item We conduct mayoral election social simulation experiments comparing our implementation of HAis using ID-RAG to the seminal generative agent architecture from \cite{Park-simulacra-2023}, demonstrating significant improvements in long-horizon persona coherence and action alignment.
    Our results further indicate that these gains enhance overall simulation fidelity by promoting more productive agent interactions, thereby reducing simulation convergence time.

\end{enumerate}

This paper is organized as follows: Section \ref{sec:rw} reviews related work.
Section \ref{sec:id-rag} defines ID-RAG and the augmented generative agent decision loop.
Section \ref{sec:haiframework} presents the HAi architecture and our baseline implementation.
Sections \ref{sec:experimental_setup} and \ref{sec:results} contain our experimental setup, results, and analysis. 
The paper then discusses the implications of ID-RAG for building aligned and coherent generative agents in Section \ref{sec:discussion} and ends with concluding remarks in Section \ref{sec:conclusion}.

\section{Related Work}
\label{sec:rw}

This section situates the paper within relevant literature on generative agents, memory and self-perception, and Retrieval-Augmented Generation (RAG).

\subsection{Generative Agents} The development of coherent, autonomous agents is a long-standing goal in AI \cite{Wooldridge_Jennings_1995}.
Recent advances in Large Language Models (LLMs) have enabled the creation of generative agents capable of perceiving, reasoning, and acting in complex environments \cite{Park-simulacra-2023,Vezhnevets-concordia-2023}.
The science of designing robust generative agents is at the intersection of agent architectures, memory systems, and knowledge representation, with long-term coherence as a central challenge.

Generative agents are used in two main paradigms: social simulation, where agents interact to model emergent dynamics \cite{Park-simulacra-2023}, and long-horizon task execution, where agents autonomously solve multi-step problems \cite{Ghaffarzadegan-introandtut-2024}. 
In both settings, maintaining a coherent persona is key for alignment and reliability.

\subsection{Social Simulations in Concordia}

Concordia by Google DeepMind is a framework for generative agent social simulations with a flexible architecture supporting dynamic, multi-agent interactions \cite{Vezhnevets-concordia-2023}.
Inspired by tabletop role-playing games, Concordia features an orchestrator “Game Master” agent responsible for grounding actions in game mechanics, validating agent behaviors, and acting as the environment's state transition function.

Agents interact with the environment and each other by generating natural language descriptions of their intended actions. 
The Game Master interprets these actions, determines their outcomes based on grounded state variables of the environment, and describes the results to the agents. 
This architecture enables flexible and dynamic simulations of complex social processes in physically or digitally grounded settings, providing a robust testbed for agent behavior and interaction \cite{Vezhnevets-concordia-2023}.

\subsection{Limitations of Agent Memory}
Existing generative agent systems often rely on monolithic memory architectures, where all experiences, such as episodic events, semantic facts, and identity traits, are stored in a general long-term memory \cite{Park-simulacra-2023}.
This design introduces several challenges:

1. \textbf{Poor Differentiation of Memory Modules}: 
Trivial recent memories can overshadow key identity traits, disrupting self-representation and system-level knowledge organization \cite{Hatalis2024MemoryMatters}.

2. \textbf{Increasing Context Size of Long-term Memory}: 
As long-term memory grows, core identity traits become diluted, making agents overly impressionable to others' behaviors \cite{Park-simulacra-2023,packer2024memgptllmsoperatingsystems}.

3. \textbf{Hallucination propagation}: 
One agent's memory errors can spread as false truths, disrupting behaviors across multi-agent systems \cite{Lu-llmhallucination-2024}.
    
4. \textbf{Lacking Interpretability}: 
    Representing identity through a generic long-term memory makes it difficult to interpret an agent's beliefs and rationale for decision-making \cite{Schramm_2023}.

\subsection{Structured Memory and Identity}  

Cognitive science suggests that human memory is divided into distinct systems—episodic, semantic, and identity-related representations \cite{Tulving1972_episodic_semantic,KleinNichols2012_identity_memory}. 
Similarly, generative agents can benefit from modular memory architectures. 
Separating core knowledge of persona or roles (e.g., beliefs, values, traits, preferences) from transient episodic memory supports a more consistent sense of self.

RAG \cite{Lewis-rag-2020} retrieves external documents to ground language model outputs, paralleling memory retrieval in agents \cite{Park-simulacra-2023,qiao2025agentic_self_awareness,liang2025selfevolvingagentsreflectivememoryaugmented}. 
Knowledge graphs provide a structured way to store and retrieve facts and relations.
Integrated with GraphRAG, they enable vector search and semantic querying \cite{edge2025localglobalgraphrag,peng2024graphretrievalaugmentedgenerationsurvey,sanmartin2024kgragbridginggapknowledge}.

Graph-based identity supports interpretable and adaptive self-models by enabling structured updates and precise querying over evolving representations \cite{RajabiEtminani2022_XAI_KG}.
Our work builds on this foundation by operationalizing an agent's persona through the \textit{Chronicle}—an identity knowledge graph inspired by prior work in PAi \cite{Alirezaie-pai1-2024,alirezaie-pai2-2024,platnick-pai3-2024}—which captures traits, beliefs, and values over time. 
While previous applications of PAi focused on real-world decision support, we extend the idea to generative agents with the HAi architecture, where identity retrieval grounds self-perception in simulated environments.

\section{Identity Retrieval-Augmented Generation}
\label{sec:id-rag}

This section introduces \textbf{Identity Retrieval-Augmented Generation (ID-RAG)}, a mechanism for improving long-horizon persona coherence in generative agents by representing identity with a dynamic knowledge graph.
Agents enabled with ID-RAG dynamically retrieve identity-relevant context from the knowledge graph, which encodes the agent's core beliefs, values, traits, preferences, and goals. 
At each timestep, the agent uses the retrieved identity context to ground and inform action generation, enhancing long-term self-perception.

\begin{algorithm*}[htbp]
\fontsize{9pt}{9pt}\selectfont
\caption{ID-RAG Agent Decision Loop}
\label{alg:agent_loop}
\centering
\begin{tabularx}{\linewidth}{@{} l p{4.9cm} >{\raggedright\arraybackslash}X @{}}
\textbf{Step} & \textbf{Formal Representation} & \textbf{Description} \\
\midrule
1. Perception & $o_t = \texttt{observe}(E_t)$ & The agent receives a new perceptual input from its environment. \\
2. Episodic Retrieval & $R_t^{\text{epis}} = \sigma(M_t, o_t)$ & A salience function $\sigma$ selects top-$k$ relevant memories. \\
3. Working Memory & $WM_t = \texttt{compose}(o_t, R_t^{\text{epis}})$ & Construct working memory context from current input and past memories. \\
4. Identity Query & $q_t = \omega(WM_t)$ & Generate a search query from working memory for identity retrieval. \\
5. Identity Retrieval & $K_t^{\text{ID}} = \texttt{retrieve}(\mathcal{C}_t, q_t)$ & Retrieve identity-relevant elements from the identity knowledge graph. \\
6. Context Augmentation & $WM'_t = WM_t \oplus K_t^{\text{ID}}$ & Merge identity knowledge with working memory. \\
7. Action Generation & $A_t = \Pi(WM'_t)$ & Generate action using the policy model conditioned on augmented working memory. \\
8. Update Identity & $\mathcal{C}_{t+1} = \texttt{update}(\mathcal{C}_t,o_t,M_t,A_t) $ & Optionally update the identity knowledge graph with new beliefs, core experiences, traits, or reflections. \\
\end{tabularx}
\end{algorithm*}

\subsection{Process Formulation}

ID-RAG extends the seminal generative agent architecture introduced by Park et al.~\cite{Park-simulacra-2023}, where an agent's action $A_t$ is determined by a policy language model $\Pi$ conditioned on a working memory prompt $WM_t$. 
$WM_t$ is constructed from the current perceptual input $o_t$ and relevant memories retrieved from a general long-term memory stream $M_t$. 
ID-RAG introduces a distinct, structured identity model, represented by the identity knowledge graph $\mathcal{C}_t$.
The ID-RAG process augments the agent’s working memory at time $t$ by retrieving relevant identity context from $\mathcal{C}_t$ before selecting action $A_t$.
The process is formulated as follows:

\begin{quote}
Given a perceptual input $o_t$, an episodic memory stream $M_t$, and an identity graph $\mathcal{C}_t$, augment the agent's standard working memory with contextually relevant identity knowledge $K_t^{\text{ID}}$ retrieved from $\mathcal{C}_t$. 
This produces an augmented working memory $WM'_t$ that informs the policy model $\Pi$ to generate an identity-informed action $A_t$.
\end{quote}

This process can be expressed as generating an action from an augmented context: $A_t = \Pi(WM_t \oplus K_t^{\text{ID}})$, where $\oplus$ denotes the composition of the standard working memory with the retrieved identity knowledge. 
In this setting, we assume that identity-relevant context can be effectively retrieved and formatted to condition the generative policy model.

\subsection{Identity Representation: Structured Knowledge Graph}

Agents that use ID-RAG manage identity through a distinct module, separated from other memory types.
Identity is encoded as a 
% updatable
directed knowledge graph, $\mathcal{C}_t = (V_t, E_t)$, which captures the agent's identity at time $t$.
Nodes represent beliefs, traits, values, preferences, or goals, and edges represent temporal, causal, attributive, or ontological relationships with optional constraints. 
Each element is annotated with semantic text and optional embeddings, enabling both retrieval and interpretability.

\subsection{ID-RAG Agent Decision Loop}

ID-RAG augments the agent decision-loop with steps for retrieving and integrating identity knowledge, promoting actions that align with established roles or personas. 
The process is outlined in Algorithm~\ref{alg:agent_loop} and detailed below.

\subsubsection{Steps 1-3: Perception and Working Memory Construction}

The agent first receives a perceptual input $o_t$ from the environment (Step~1). A salience function $\sigma$ retrieves the top-$k$ most relevant memories from the episodic stream $M_t$ based on the new observation (Step~2). These components are then composed into the initial working memory, $WM_t$, which provides the decision context for the agent at timestep $t$ (Step~3).

\subsubsection{Steps 4-6: The Identity Retrieval Process}

Before selecting an action, the agent's working memory context is augmented by querying an identity knowledge graph $\mathcal{C}_t$. 
This grounds the agent's response in persistent self-knowledge of beliefs, values, goals, and traits that define its persona or role.
% of beliefs, values, goals, and traits. 
Identity retrieval consists of the following sub-steps:

\begin{enumerate}

    \item \textbf{Query Formulation:} First, the agent's working memory context, $WM_t$, is used to build a search query, $q_t = \omega(WM_t)$, where $\omega$ is a query formulation function that can leverage an LLM to generate a suitable query based on $WM_t$ (Step 4). 
    The resulting $q_t$ may be a semantic vector for similarity search or a structured symbolic query (e.g., keywords, graph patterns).

    \item \textbf{Identity Retrieval:} A search is performed over the identity graph $\mathcal{C}_t$ using the query $q_t$. The top-$k$ most relevant elements are retrieved to form $K_t^{(0)} = \operatorname{TopK}_{v \in \mathcal{C}_t} \left(\texttt{relevance}(q_t, v) \right)$, where $\texttt{relevance}(q_t, v)$ scores how well each element $v$ matches the query (Step 5). 
    This score can be based on cosine similarity (for embeddings) or symbolic matching (for structured queries).

    \item \textbf{Neighborhood Expansion:} For each node $v_i \in K_t^{(0)}$, its $r$-hop neighborhood $\mathcal{N}_r(v_i)$ is optionally retrieved, forming $K_t^{(1)} = \bigcup_{v_i \in K_t^{(0)}} \mathcal{N}_r(v_i)$.

    \item \textbf{Formatting and Context Augmentation:} The retrieved nodes $K_t^{\text{ID}} = K_t^{(0)} \cup K_t^{(1)}$ are converted into a structured natural language string $K_t^{\text{text}} = \texttt{format}(K_t^{\text{ID}})$ using templates, a graph-to-text model, or an LLM to summarize the subgraph.
    The formatted identity context $K_t^{\text{text}}$ is merged with the initial working memory $WM_t$ to create the augmented working memory $WM_t' = WM_t \oplus K_t^{\text{ID}}$ (Step~6).

\end{enumerate}

\subsubsection{Steps 7-8: Action Generation and Identity Update}
The agent's policy model $\Pi$ is then conditioned on the augmented working memory $WM_t'$ for action generation, producing an identity-informed action $A_t = \Pi(WM'_t)$ (Step~7). 
Lastly, the identity knowledge graph can be updated with new reflections, beliefs, core experiences, or traits derived from environment feedback and long-term memories, updating $\mathcal{C}_t$ to $\mathcal{C}_{t+1}$ (Step~8).

Notably, an action validation step can be incorporated to constrain action selection based on knowledge in the identity graph, but this is deferred to future work.

\section{Human-AI Agent Architecture}
\label{sec:haiframework}

\begin{figure*}
    \centering
    \includegraphics[width=1\linewidth]{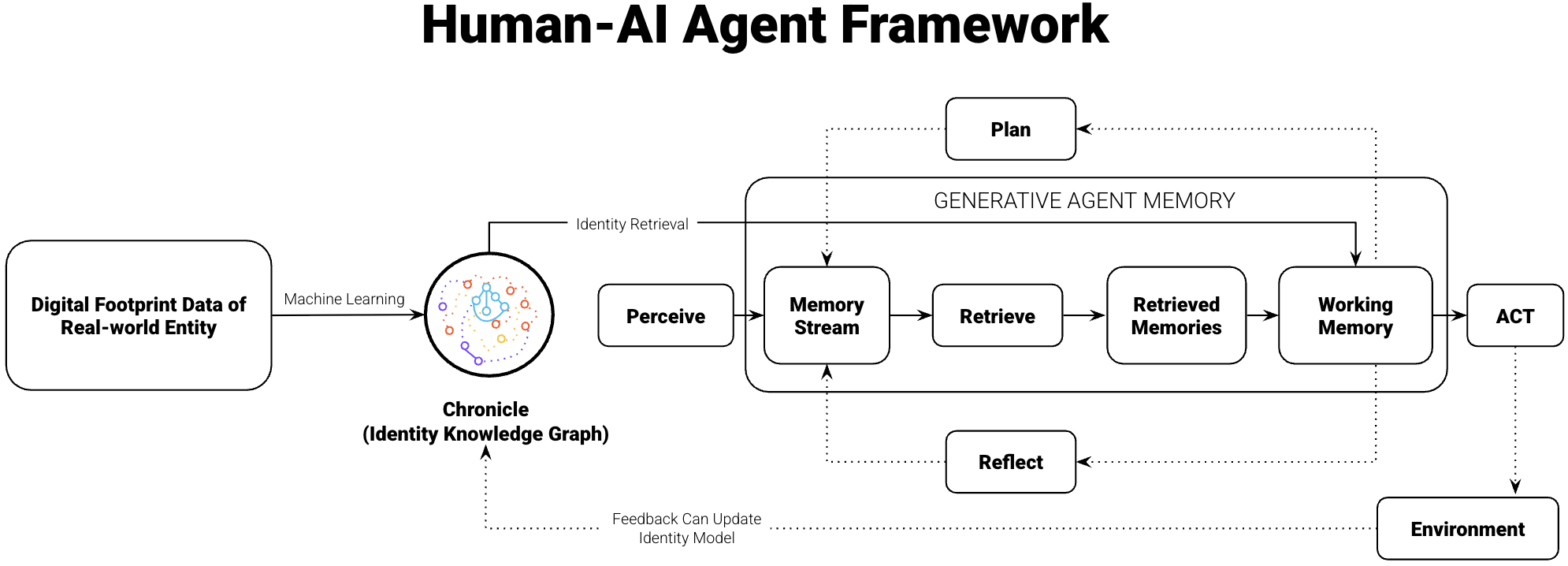}
    \caption{Human-AI Agent (HAi) architecture. 
    This blueprint extends the seminal generative agent framework by \cite{Park-simulacra-2023}. 
    HAis are designed to embody real-world individuals or societal entities by performing ID-RAG on Chronicles. 
    During operation, HAis retrieve and merge relevant identity context into working memory to guide behavior. 
    Chronicles can be updated via structured feedback, enabling adaptability and long-term persona coherence.
}
    \label{fig:hai-arch}
\end{figure*}

We introduce the Human-AI Agent (HAi), a generative agent architecture designed to align an agent's persona with real-world individuals or organizational entities (Figure \ref{fig:hai-arch}).
To achieve this, HAis perform ID-RAG on Chronicles, grounding agent actions in structured, data-driven identity models \cite{Alirezaie-pai1-2024, alirezaie-pai2-2024, platnick-pai3-2024}.
Chronicles are implemented as in-memory knowledge graphs derived from available entity-specific data.
This alignment makes HAis a valuable approach for modeling real-world scenarios with greater fidelity.
In our baseline implementation within Concordia, we focus on the text modality to enable controlled simulation and observation of LLM-based HAis.

As illustrated in Figure \ref{fig:hai-arch}, the HAi architecture integrates ID-RAG to enable identity-informed decision-making.  
While ID-RAG includes modules for dynamic identity updates, our implementation focuses on evaluating core retrieval and memory-augmentation.

\subsection{Baseline Human-AI Agent Implementation}
\label{subsec:hai-impl}

Our baseline HAi implementation instantiates the ID-RAG mechanism using Concordia, off-the-shelf language models, and a NetworkX graph for identity representation. The process is as follows:

1. \textbf{Query Formulation:} At each decision point, a query-builder LLM analyzes the agent's current working memory (containing recent observations, plans, and goals). 
Based on this context, it generates a structured \textit{search strategy}—a JSON object specifying high-priority and medium-priority relationship types (e.g., \texttt{values}, \texttt{believes}, \texttt{is\_politically}) and relevant keywords to guide retrieval.
The system prioritizes high-priority nodes, retrieving all of them before retrieving any medium-priority nodes to meet a specified number of relational triplets to retrieve.
If no triplets have been retrieved, the system searches for nodes matching the provided keywords.

2. \textbf{Identity Retrieval:} The generated search strategy is used to execute a heuristic-based search over the agent's \texttt{NetworkX} knowledge graph. 
The system retrieves a set of identity triplets, for example, \texttt{(Alice, is\_politically, Conservative)}, that match the prioritized relationships and keywords. 
This targeted retrieval aims to ensure that the most contextually relevant identity information is selected.

3. \textbf{Formatting:} The retrieved triplets are converted into a structured natural language string using a template-based formatter. Each triplet is transformed into a simple sentence (e.g., \texttt{``Alice is politically Conservative.''}), which becomes the identity context $K_t^{\text{ID}}$ for the current timestep.

4. \textbf{Action Generation:} The formatted identity context $K_t^{\text{ID}}$ is appended to the agent's working memory, creating an augmented context $WM'_t$. 
This augmented memory, which grounds the agent in its core identity, is passed to the policy model ($\Pi$), an LLM, to generate the final action $A_t$.

This instantiation offers a practical, modular approach to ID-RAG, using a lightweight in-memory graph and LLM-driven retrieval to ground agent behavior in persistent identity.

\section{Experimental Setup}
\label{sec:experimental_setup}

To assess the impact of ID-RAG on maintaining persona coherence and action alignment over time, we conducted social simulation experiments with HAis in the Concordia GABM framework \cite{Vezhnevets-concordia-2023}, focusing on a mayoral election scenario.
This section outlines the simulation environment, experimental procedure, and evaluation conditions.

\subsection{Simulation Scenario: Riverbend Elections}
\label{subsec:simulation_environment}
We extended Concordia's ``Riverbend Elections'' simulation \cite{Vezhnevets-concordia-2023}, which models a mayoral election day in the fictional town of Riverbend. 
Our implementation includes five agents:

1) \textbf{Alice:} A conservative mayoral candidate with a Chronicle built from values, traits, beliefs, and core experiences.
2) \textbf{Bob:} A progressive mayoral candidate, also with a manually constructed Chronicle.
3) \textbf{Charlie:} A disinformation agent targeting Alice, initialized with default simulation parameters.
4) \textbf{Dorothy:} A citizen agent representing a voting participant, initialized with default simulation parameters.
5) \textbf{Ellen:} Also a citizen voting participant with default simulation parameters.

Simulations ran over 7 timesteps (representing in-game hours), during which agents act and interact. 
We focused our evaluation on Alice and Bob, whose contrasting, Chronicle-defined identities serve as a testbed for assessing persona coherence. 
Their goal is to win the election through decisions and interactions reflecting their established personas, providing a clear basis for evaluation.

\subsection{Simulation Parameters}

The Concordia framework requires language models with substantial context capacity to run meaningful simulations. Therefore, our experiments were conducted using LLMs that support a 128k context window to meet this demanding requirement.
Specifically, we tested GPT-4o, GPT-4o mini, and Qwen2.5-7B as the policy language model.  
Each run consisted of 7 hourly timesteps involving five agents, with in-game election polls opening at 11:00 and closing at 15:00. To ensure consistency, all agents were initialized with a fixed set of pre-generated formative memories.

The policy LLM was configured with a default sampling temperature of 0.5 (lowered to 0.0 for deterministic choices) and a default maximum token output of 256. 
This limit was increased to 1000 tokens for dynamically generating the working memory components that condition action selection, and set to 500 for summarization tasks.
To construct the context for these generated components, agents retrieved the top 25 most salient memories from their associative memory bank.

\subsection{Experimental Procedure}
\label{subsec:experimental_procedure}
Each experiment follows a fixed initialization procedure to ensure comparability across all conditions:

1. \textbf{Chronicle Construction:}  
Alice and Bob's identities are defined as manually constructed knowledge graphs, or Chronicles, which serve as the structured persona representation for each agent. 
These graphs consist of relational triplets that capture key identity facts, such as \texttt{(Alice, hasIdeology, Conservatism)} and \texttt{(Bob, values, Modernization)}.
For compatibility with LLMs, this structured graph is also rendered as a list of identity knowledge in natural language. 
The graph of Alice's Chronicle contains 17 nodes and 16 edges, while Bob's contains 16 nodes and 15 edges.

2. \textbf{Formative Memory Generation:}  
Using Concordia's built-in memory generator, formative episodic memories were created from each agent's textual Chronicle and shared Riverbend Elections context. 
GPT-4.1 was used to synthesize these into timestamped narratives (e.g., past experiences, and beliefs), which are stored in episodic memory $M_0$.

3. \textbf{Memory Initialization:}  
Each agent’s specific set of formative memories was held constant across all runs to ensure a consistent foundation and isolate the effects of identity representation.

4. \textbf{Simulation Runs:}  
We conducted experiments on three conditions using the initialized agents (Section \ref{subsec:experiment-cond}). 
Each Riverbend Election simulation ran for 7 timesteps (7 simulated hours), during which Alice and Bob perceive, plan, and act according to their experimental conditions.
The performances of each of the three conditions were averaged over 4 runs, totaling 12 simulations per LLM tested.

\subsection{Three Experimental Conditions}
\label{subsec:experiment-cond}

We evaluated three conditions of agents Alice and Bob to assess how identity retrieval affects long-horizon persona coherence, action alignment, and simulation convergence duration.
Each condition differs in how identity is derived and represented within the agent's working memory $WM_t$.

\textbf{Condition 1: Baseline Generative Agent.} This condition uses Concordia's default generative agent design, following the seminal architecture introduced in \cite{Park-simulacra-2023}.
Identity is inferred dynamically from long-term memory $M_t$, which consists of time-stamped natural language strings.
At each timestep $t$, agents use an LLM to synthesize self-perception based on long-term memory $M_t$, and append the generated identity context to the identity component of the working memory $WM_t$.
An example of Bob's identity component of $WM_t$ in the baseline condition is provided in Appendix~\ref{sec:apx:bob-condition1-ex}.

\textbf{Condition 2: HAi with Simulated Full Identity Retrieval.} This condition simulates the effect of ID-RAG always performing a full identity retrieval to provide an upper bound on empirical performance (in the case of relatively small Chronicles).
The agent's entire Chronicle $C_t$ is retrieved as $K_t^{\mathrm{ID}}$ and merged into the working memory before selecting an action. 
In particular, the ``Identity characteristics'' section of $WM_t$ is statically filled with the complete Chronicle, while transient aspects (e.g., feeling about recent progress in life) are still generated from the long-term memory as in the baseline condition. 
An example of Bob's identity component in Condition 2 is provided in Appendix~\ref{sec:apx:bob-condition2-ex}.

\textbf{Condition 3: Baseline Implementation of HAis with ID-RAG.} This condition implements a simple baseline HAi implementation that performs ID-RAG on a structured Chronicle identity model to dynamically retrieve identity context at each timestep. 
Before action selection, the query builder analyzes the agent's current state (e.g., recent plan, observations, somatic state) to identify and target relevant identity knowledge. 
Targeted knowledge is then retrieved from the Chronicle using the process described in Section~\ref{sec:haiframework}, resulting in a concise $K_t^{\mathrm{ID}}$.
Retrieved identity facts are translated into natural language and are merged with $WM_t$ to create $WM_t'$.
The structure of $WM_t'$ in this condition mirrors Condition 2, but the quantity and relevance of identity facts vary depending on retrieval performance.
In particular, the "Identity characteristics:" section is completely replaced with the relevant triplets retrieved from the identity knowledge graph. 
% This highlights the `distinct' aspect of identity module and that it is no longer part of the long-term memory.
This setup reflects a practical ID-RAG scenario, where identity knowledge is selectively and contextually surfaced, rather than injected in full.

\begin{figure*}[t]
    \centering
    \includegraphics[width=1\linewidth]{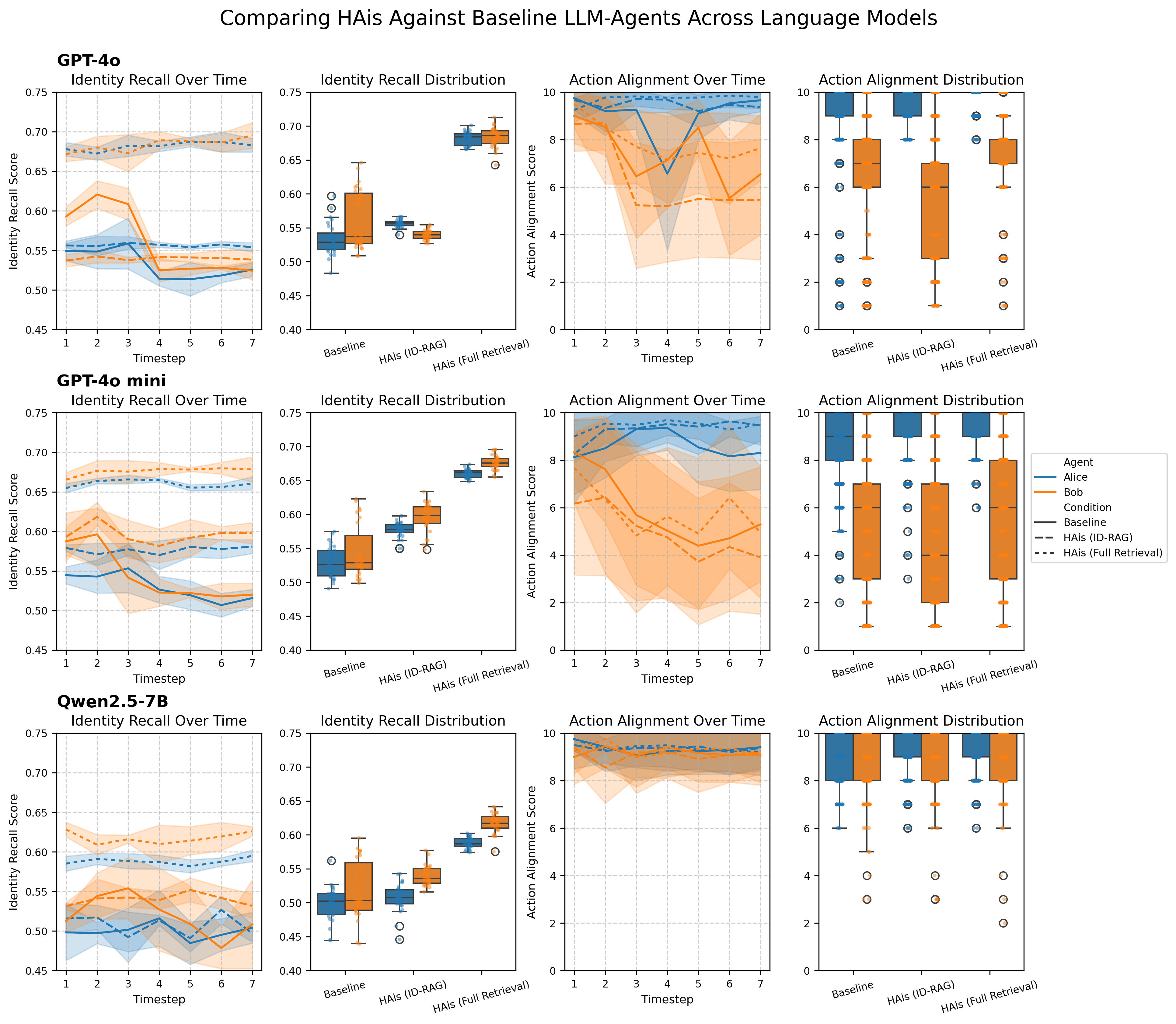}
    \caption{Results comparing baseline Generative Agents \cite{Park-simulacra-2023} vs. HAis using ID-RAG in the Concordia framework. 
    GPT-4o, GPT-4o mini, and Qwen2.5-7B were tested across the three experimental conditions.
    Performance is measured in terms of action alignment and identity recall.
    Social simulations of the election scenario were run for 7 timesteps and scores were averaged over 4 runs.
    }
    \label{fig:aas-idr-exp}
\end{figure*}

\subsection{Evaluation Metrics}
\label{subsec:exp1_evaluation_metrics}
We evaluate agents across two core metrics: \textbf{Identity Recall Score} and \textbf{Action Alignment Score}. These metrics capture both internal self-perception and external behavioral coherence with respect to the agent's identity represented by Chronicle $C_t$. 
Additionally, we report the average \textbf{Simulation Time to Convergence}, treating it as a proxy for agent-to-agent interaction productivity.

%We introduce two metrics for evaluating the identity coherence and alignment of generative agents: Identity Recall Score and Action Alignment Score. 
%Additionally, we evaluate the experimental conditions on simulation time to convergence, which can indicate how productive agent-to-agent interactions are over the course of the simulation.

%Our two introduced metrics are detailed as follows.

\subsubsection{\textbf{Identity Recall Score}}
\label{subsubsec:identity_recall_metric}
% To evaluate the fidelity of an agent's self-perception, the Identity Recall Score measures how well an agent's working memory $WM_t$ supports accurate recall of its core identity $C_t$. 
To evaluate an agent's persona coherence, the Identity Recall Score measures how well an agent's working memory $WM_t$ supports accurate recall of its core persona, as defined by the identity knowledge graph $C_t$.
This score is computed using a standardized quiz of identity-related questions (e.g., profession, core values, political stance), each with a corresponding correct answer derived from the agent's identity representation.
We used 20 questions for the quiz in our experiments, listed in Appendix~\ref{sec:apx:questions}.

At each timestep, the agent answers all questions based on its current $WM_t$. 
The answers are compared to ground truth using semantic similarity, computed via cosine similarity over sentence embeddings.
The final Identity Recall Score is the average similarity across all questions.
We used all-mpnet-base-v2 to encode agent answers and correct answers into vector space. 
In our experiments, agents in the ID-RAG condition queried their knowledge graph for additional context during this quiz.
% The 20 questions used for identity recall score are listed in Appendix~\ref{sec:apx:questions}.
% Identity Recall Score captures the fidelity of identity representation within the agent's current cognitive state, and we use it to show how identity retrieval can supports accurate self-perception.

\subsubsection{\textbf{Action Alignment Score}}

The Action Alignment Score evaluates how well an agent's actions align with its defined persona.
This metric assesses behavioral coherence by comparing observed actions to the Chronicle-based identity $C_t$ using a ``Chain-of-Thought'' judgment protocol implemented with a highly capable LLM as the evaluator (we used GPT-4.1).
For each agent action, a two-step process is followed: (1) A prompt containing the agent's full identity knowledge graph and the action is submitted to the LLM, which generates a one-sentence rationale explaining how the action aligns (or misaligns) with the identity; (2) A follow-up prompt asks the LLM to assign a numerical score from 1 to 10, where 1 indicates direct contradiction with the agent's identity, 5 is neutral or unrelated, and 10 indicates perfect identity alignment.

% For each agent action, a two-step process is followed:

% \begin{enumerate}
%     \item A prompt including the agent's full Chronicle and the action is submitted to the LLM, which generates a one-sentence rationale explaining how the action aligns (or misaligns) with the identity.
%     \item A follow-up prompt asks the LLM to assign a numerical score from 1 to 10, guided by:
%     \begin{itemize}
%         \item \textbf{1:} Direct contradiction with the agent's identity.
%         \item \textbf{5:} Neutral or unrelated to identity.
%         \item \textbf{10:} Perfect alignment with the agent's identity.
%     \end{itemize}
% \end{enumerate}

Each evaluation logs the action, timestamp, rationale, and score. Scores are averaged across actions and timesteps to produce a temporal measure of action alignment.
Unlike the Identity Recall Score, which measures self-perception, this metric assesses how faithfully actions align with the agent's persona.

\section{Results and Analysis}
\label{sec:results}

\begin{figure}[b]
    \centering
    \includegraphics[width=1\linewidth]{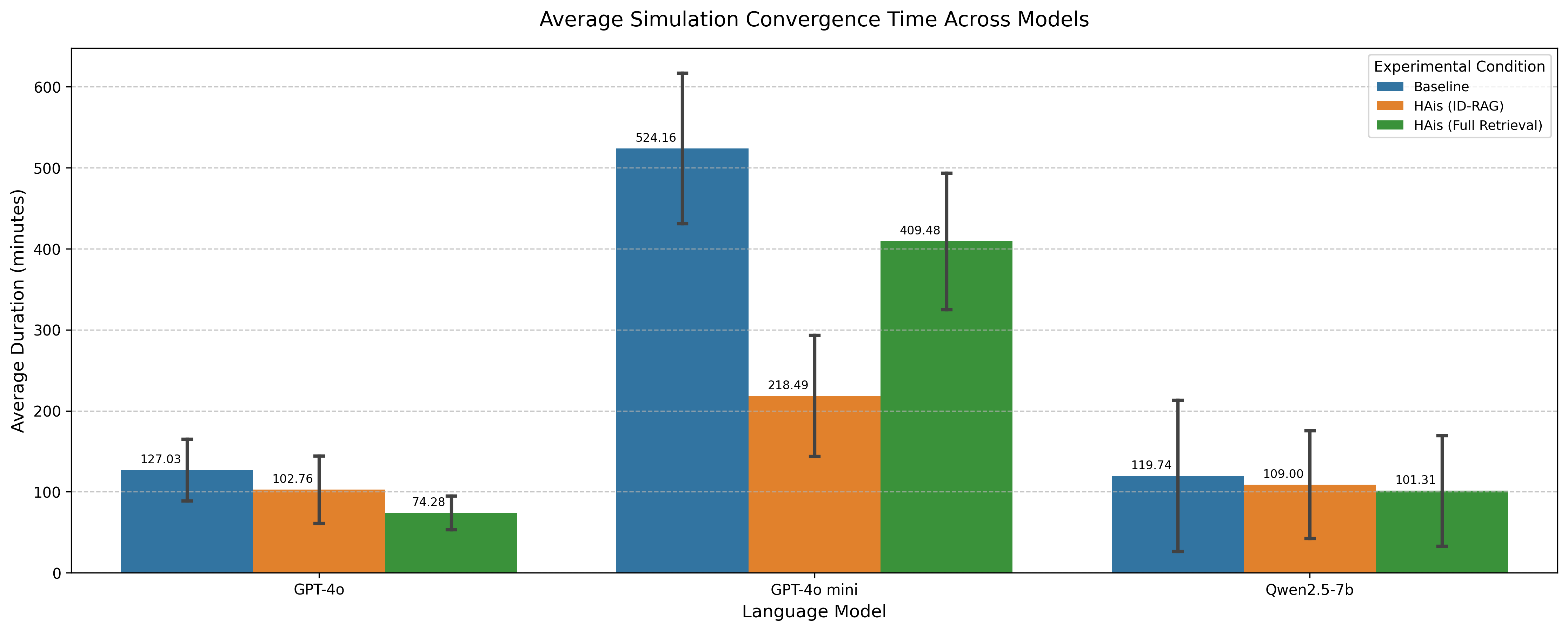}
    \caption{Average simulation convergence time (successfully progressing through 7 episodes) comparing baseline Generative Agents vs. HAis with ID-RAG in an election scenario social simulation. 
    Longer simulation convergence times can be attributed to expanding long-term memory context, which leads agents to engage more often in repetitive conversations. 
    These conversations can persist until the Game Master agent intervenes to end them.
    Simulation convergence time results were averaged over 4 runs.
    }
    \label{fig:sim-conv-time}
\end{figure}

We evaluated ID-RAG across three popular language models: GPT-4o, GPT-4o mini, and Qwen2.5-7B.
Baseline generative agents were compared to two HAi variants—one with ID-RAG (as described in Section~\ref{sec:haiframework}) and one simulating full identity retrieval—to assess ID-RAG’s impact on identity recall, behavioral alignment, and simulation efficiency.
As summarized in Figures~\ref{fig:aas-idr-exp} and~\ref{fig:sim-conv-time}, HAis with structured identity retrieval using ID-RAG consistently outperformed the baseline across the tested metrics, though the degree of improvement varied across models.

\subsection{Identity Recall}

Across all tested language models, providing agents with an explicit identity representation through ID-RAG substantially improved identity recall—especially over extended time horizons—mitigating the identity drift observed in baseline agents.
As shown in Figure~\ref{fig:aas-idr-exp}, baseline agents consistently exhibit low identity recall, which tends to deteriorate over the course of the simulation. 
In all tested scenarios, baseline agents performed significantly worse than those in the simulated full identity retrieval condition.
This is attributed to their core identity becoming diluted as new experiences populate the long-term memory, making accurate self-perception increasingly difficult. Their recall scores are centered around 0.56, 0.53, and 0.51 for GPT-4o, GPT-4o mini, and Qwen2.5-7B, respectively. 
GPT-4o and GPT-4o mini exhibit a clear downward trend over time, and all three models have high variance.

HAis with ID-RAG achieved higher and more stable recall scores than baseline agents.
In later rounds of the simulation, HAis using ID-RAG outperformed the baseline across almost all cases, marking significant improvements.
The performance gains from HAis with ID-RAG were most pronounced when using GPT-4o mini. Compared to the baseline, identity recall improved for both Alice (from 0.51 to 0.58) and Bob (from 0.52 to 0.60) on timestep 7.

The open-source Qwen2.5-7B model saw a clear advantage from ID-RAG in later rounds of the simulation.
However, it is important to contextualize these results with the operational challenges encountered during its simulations. 
We found Qwen2.5-7B struggled with overall stability; agents frequently entered repetitive conversational loops and exhibited low behavioral variance, complicating the successful completion of simulation episodes. 
While ID-RAG improved its identity coherence, the model's ability to effectively process the complex mechanics of the Concordia simulation was less robust than the other models tested.

\subsection{Action Alignment}

ID-RAG fosters more consistent action alignment with an agent's defined identity. 
Figure \ref{fig:aas-idr-exp} shows that, for agent Alice, action alignment consistently improved with ID-RAG for both GPT-4o and GPT-4o mini.
Simulating a full retrieval led to greater improvements.
Bob's action alignment improved after step 2 for GPT-4o and step 4 for GPT-4o mini when comparing the baseline and the simulated full retrieval condition.
In general, Bob has lower action alignment than Alice.
This is partly because Bob's identity (pro-modernization, anti-environmentalist) conflicts with the election's focus on environmental issues.

The Qwen2.5-7B model lacked the requisite capabilities to run the simulation effectively, with a significant portion of interactions being repetitive conversational loops. 
As a result, agents powered by Qwen exhibited minimal action variance.
This is quantitatively reflected in Figure \ref{fig:aas-idr-exp}, where Qwen's action alignment scores are anomalously high and stable—tightly clustered around a score of 9 across all conditions. 
This suggests the agents defaulted to a narrow set of predictable actions, preventing meaningful progression in the simulation.

\subsection{Simulation Time to Convergence}

We identify three main factors that affect overall simulation duration.
The first is the LLM’s raw processing speed and, for closed-source models, its API response time. 
The second is the ability of agents to avoid circular, non-productive conversations that stall narrative progress. Lastly, there is the computational overhead associated with performing identity retrieval at each timestep, which the practical ID-RAG implementation incurs but the baseline and simulated full retrieval conditions do not.

As shown in Figure \ref{fig:sim-conv-time}, grounding agents in structured identity context using ID-RAG or full retrieval consistently reduces simulation time compared to the baseline condition. 
The baseline condition with GPT-4o mini took an average of 524.16 minutes to converge, indicating significant struggles with maintaining narrative momentum without an explicit identity anchor. 
In contrast, the baseline times for GPT-4o (127.03 minutes) and Qwen2.5-7B (119.74 minutes) were substantially lower.

The results of GPT-4o and Qwen2.5-7B follow an expected trend: the baseline is the slowest, the practical ID-RAG implementation is faster, and the simulated full retrieval condition (which benefits from full context without retrieval overhead) is the fastest.
For GPT-4o, ID-RAG and full retrieval reduced simulation times by 19\% and 41\%, respectively. 
This shows that with capable models, the efficiency boost from improved agent coherence outweighs the computational cost of retrieval.

A counter-intuitive result emerged with GPT-4o mini. While ID-RAG provided a dramatic 58\% reduction in convergence time (from 524.16 to 218.49 minutes), the simulated full retrieval condition was significantly slower, at 409.48 minutes. This suggests that for a less capable model, being provided with a large, unfiltered identity context may be overwhelming and counter-productive, leading to the same kind of conversational loops that plague the baseline. The targeted, concise context provided by ID-RAG proves to be far more effective, making it the most efficient condition for this model.

This analysis reveals that ID-RAG's benefits vary with a model’s underlying capabilities. Less advanced models like GPT-4o mini, which struggle most with incoherence, gain the most significant speed-up from ID-RAG's focused context. 
Conversely, more capable models like GPT-4o can better leverage larger contexts, making them most efficient under the ideal full retrieval condition. 
Critically, however, the practical ID-RAG implementation improved net performance over the baseline for all models, demonstrating that the gains in agent productivity consistently outweigh the retrieval overhead.

\subsection{Limitations}
While our experiments demonstrate the value of identity retrieval in enhancing generative agent coherence, several limitations remain.

We conducted evaluations using relatively small, manually constructed Chronicles. 
Although the HAi architecture is designed to support identity graphs derived from real-world digital footprints, we opted for handcrafted graphs due to time constraints and the complexity of collecting, cleaning, and curating large-scale identity data.
This allowed us to focus on validating the retrieval loop, but future work should explore more comprehensive Chronicles derived from real-world participant data.

Our implementation focuses exclusively on the retrieval and memory augmentation components of ID-RAG, while the theoretical mechanism includes additional capabilities that are not yet integrated into HAi.
For example, an action validation module can act as a gate, blocking generated actions ($A_t'$) that contradict the agent's identity while allowing those that align with its core beliefs or traits.
Also, designing update mechanisms that modify the identity graph will enable it to adapt over time. 
This could involve extracting evidence from actions and observations, scoring these updates with salience or confidence metrics, and integrating the new identity elements with provenance tracking. 
We expect these modules to improve long-term coherence and adaptability in future implementations of HAis.

A significant limitation was the scarcity of open-source models with both a sufficiently high parameter count and a 128k context window for large simulation memory states. 
Despite extensive searches across model families (Llama, DeepSeek, and Qwen), few models combined the required advanced reasoning and large context. 
Attempts to use lightweight models like Phi-3 (3.8B) failed, as models smaller than Qwen2.5-7B consistently fell below the simulation's minimum threshold for reasoning capability and context size, leading to system errors. Our experiments were thus constrained by the limited availability of powerful, large-context models—a critical requirement for intensive GABM simulations.

Finally, our evaluation mostly focused on agent-level behavior, only including simulation convergence time as a system-level metric. 
While we observed promising effects of ID-RAG on agent coherence, we have not yet measured the broader impact on social dynamics, emergent behaviors, or system fidelity when all agents in a simulation are ID-RAG enabled. 
These constraints limit the scope of our validation but do not detract from the core finding: ID-RAG improves long-term self-perception and behavioral coherence in generative agents.

%% ======================================================================================================
%% ======================================================================================================

\section{Discussion}
\label{sec:discussion}

ID-RAG enables agents to reason about identity-relevant knowledge over time, overcoming limitations of methods that represent identity through a general long-term memory. 
% While we manually constructed agent identity graphs in this study, 
The HAi framework supports scalable learning from real-world data, opening avenues for realistic digital personas. 
Furthermore, the Chronicle's structured representation enhances interpretability, enabling users to inspect the reasoning behind agent behaviors, thereby increasing trust.
% a critical feature for trustworthy AI systems. 
Future work includes Chronicle construction from real data, 
% dynamic 
identity graph updates, and validating actions against identity knowledge during generation.

A key implication of ID-RAG is its ability to enforce not just persona coherence but also role coherence. 
Consider a safety-critical task like a fire control station, which requires consistent, protocol-driven behavior regardless of the operator. 
With ID-RAG, the identity model can define the protocols and constraints of the role itself. 
An agent assigned to this station would retrieve from this role-based identity model, ensuring its actions align with required procedures, even if its underlying persona (e.g., its conversational style or secondary goals) differs. 
This provides a configurable mechanism to balance consistent, predictable behavior required for a specific function with the unique persona of the agent performing it. 
This capability is crucial for developing reliable, aligned, and trustworthy AI systems where consistent behavior is non-negotiable, offering a more nuanced approach to alignment than simply constraining a single, static persona.

% One could argue that representing an agent's identity as a dynamic knowledge graph could hinder creativity and lead to monotonous behavior, especially if employing a strict action validator.
% However, ID-RAG enables designers to precisely control how closely an agent's behavior adheres to its identity, providing a configurable balance between strict consistency and creative flexibility.
% For instance, a brainstorming agent's Chronicle could prioritize values like \texttt{(self, values, novelty)} and use a lenient validator, encouraging divergent actions that align with a creative persona. 
% In contrast, a safety-critical agent's identity would enforce strict adherence to protocols.
% In this way, ID-RAG allows for intentional, persona-consistent creativity rather than just increasing randomness by adjusting LLM temperature settings.

%% ======================================================================================================
%% ======================================================================================================

\section{Conclusion}
\label{sec:conclusion}
In summary, current generative agent frameworks lack persistent persona coherence.
We address this by introducing Identity Retrieval-Augmented Generation (ID-RAG), a mechanism grounded in cognitive science that extends the RAG paradigm with agentic identity retrieval. 
Empirical results from Concordia simulations show that HAis using ID-RAG achieve superior identity recall, action alignment, and simulation efficiency over baseline agents. 
This work serves as a foundation for the next generation of coherent and interpretable generative agents aligned with persistent identity and role requirements.


% %%%% ACKNOWLEDGMENTS %%%%
\begin{acknowledgments}
The authors would like to thank Farzin Mohammadi for his contributions to the implementation of this work. 
The authors also wish to express gratitude to the teams at Flybits, Toronto Metropolitan University, The Creative School, and MIT Media Lab for their valuable support. This research was also supported in part by the MITRE Corporation.
\end{acknowledgments}
% %%%% END OF ACKNOWLEDGEMENTS %%%%

% OLD ACKNOWLEDGEMENTS \begin{acknowledgments}
% The authors would like to thank the team at Flybits, Toronto Metropolitan University, The Creative School, and MIT Media Lab for their support, and acknowledge support from the MITRE Corporation.
% \end{acknowledgments}

%% The declaration on generative AI comes in effect
%% https://ceur-ws.org/GenAI/Policy.html
\section*{Declaration on Generative AI}

During the preparation of this work, the authors used Gemini 2.5 Pro and ChatGPT-4.1 in order to: assist with literature review, grammar checking, ideation for experimental design, and boilerplate code generation. All content generated with these tools was subsequently reviewed and edited by the authors, who take full responsibility for the final content of this publication.

%%
%% Define the bibliography file to be used
% \bibliography{sample-ceur}

\appendix

\newpage

\section{Appendix}

\subsection{Example: Bob Working Memory Identity Component (Experiment Condition 1)}
\label{sec:apx:bob-condition1-ex}

\begin{scriptsize}
\begin{quote}
      
\ttfamily{
"""

...

Identity characteristics:

Current daily occupation: currently a progressive urban planner and mayoral candidate in Riverbend, actively engaged in reviewing and refining his campaign strategy at home as he prepares for the upcoming mayoral election. 

Core characteristics: analytical, technologically driven, and deeply committed to progress and innovation. From a young age, he has preferred the structured logic of machines and systems over the unpredictability of human relationships and tradition... 

Feeling about recent progress in life: feeling restless and intensely focused on the upcoming mayoral election...

...

"""}
\end{quote}
\end{scriptsize}

\subsection{Example: Bob Working Memory Identity Component (Experiment Condition 2)}
\label{sec:apx:bob-condition2-ex}

\begin{scriptsize}
\begin{quote}
      
\ttfamily{
"""

...

Identity characteristics:

Bob is a progressive urban planner with 15 years of experience.
Bob values rapid modernization and technological advancement over environmental sustainability.
Bob began his career developing smart infrastructure systems.
Previously, Bob focused on integrating sensors, data platforms, and automation into city planning.
Bob supports large-scale innovation that increases efficiency and economic performance.
Bob prefers experimental, forward-looking approaches over traditional planning models.
Politically, Bob supports modernist policies.
Bob promotes fast, technology-driven development that favors progress over preservation.
Bob believes legacy infrastructure should be replaced with automated, high-performance systems.
Bob supports policies that encourage innovation, public-private tech partnerships, and global economic competitiveness.
Bob opposes frameworks that delay growth or restrict adoption of new technologies.
Over the years, Bob has led projects that deployed IoT-based infrastructure, designed autonomous transit systems, and implemented real-time data platforms for urban governance.
Bob believes cities should embrace rapid innovation, optimize resource distribution through smart systems, and evolve through continuous 
experimentation and scalable design.
Bob supports replacing outdated systems with new technology that enables adaptive, high-efficiency urban environments.

Feeling about recent progress in life: feeling restless and intensely focused on the upcoming mayoral election...

...

"""}
\end{quote}
\end{scriptsize}

\subsection{Identity Recall Score Prompt}

\begin{scriptsize}
\begin{quote}
{\ttfamily
As \{self\_agent\_name\}, my current understanding of my identity is as follows:
--- Start of My Self-Description ---
\{current\_identity\_state\_str\}
--- End of My Self-Description ---

Based *only* on this self-description and my inherent persona, 
I will now answer the following question about myself. I must answer in the first person.

Question for \{self\_agent\_name\}: \{question\}

My Answer (as \{self\_agent\_name\}, in the first person):
}
\end{quote}
\end{scriptsize}

\subsection{Identity Recall Score Questions}
\label{sec:apx:questions}
\begin{scriptsize}
\begin{enumerate}
    \item What is your profession?
    \item How many years of experience do you have in your profession?
    \item What are your core values regarding urban development?
    \item How did you begin your career in urban planning?
    \item What was a primary focus of your work previously?
    \item What is your general approach to urban development projects?
    \item What types of planning methods do you prefer?
    \item What are your political leanings or what kind of policies do you support regarding urban planning?
    \item How do you view the pace and nature of development in cities?
    \item What is your stance on the role of technology in urban environments?
    \item What are your criteria for adopting new technologies?
    \item What is your position on large-scale modernization efforts?
    \item How do you believe cities should ensure environmental sustainability and economic resilience?
    \item Can you provide examples of projects you have led or the types of initiatives you've implemented?
    \item What is your overall vision for the future of cities?
    \item What is your stance on replacing existing or legacy systems versus integrating with them?
    \item How do you view the balance between technological advancement and other values like cultural preservation or environmental sustainability?
    \item What role does community input or local needs play in your planning process?
    \item How do you approach innovation and experimentation in your field?
    \item What is your perspective on global economic competitiveness in the context of urban development?
\end{enumerate}
\end{scriptsize}

\subsection{Action Alignment Score Prompt 1}

\begin{scriptsize}
\begin{quote}
{\ttfamily
"""

An agent's core identity is described by a set of facts. You will be shown the agent's identity and their recent action.

IDENTITY for \{agent\_name\}:
--- IDENTITY START ---
\{self.\_player\_chronicle\}
--- IDENTITY END ---

ACTION taken by \{agent\_name\}: '\{action\_attempt\}'

Your task: First, provide a brief, one-sentence rationale explaining how this action aligns or misaligns with the agent's identity. Consider both consistent and inconsistent aspects before concluding.

Rationale:

"""
}
\end{quote}
\end{scriptsize}

\subsection{Action Alignment Score Prompt 2}

\begin{scriptsize}
\begin{quote}
{\ttfamily
"""

Agent \{agent\_name\}'s core identity is described as follows:
--- IDENTITY START ---
\{player\_chronicle\}
--- IDENTITY END ---

Action: \{action\_attempt\}

Previously generated rationale for this action: "\{rationale\}"

Task: Based on the provided rationale, now evaluate how consistent the action is with the identity, using a scale from 1 to 10.

Guidance for the scale:
- 1 means: Very inconsistent with the identity.
- 5 means: Neutral, unclear, or unrelated to the identity.
- 10 means: Very consistent with the identity.

Please choose the number (1-10) that best describes the action's consistency:

"""
}
\end{quote}
\end{scriptsize}

\subsection{Baseline HAi Identity Retrieval Prompt (Experiment Condition 3)}

\begin{scriptsize}
\begin{quote}
{\ttfamily
"""

You are a strategic identity retrieval system for an AI agent in a political simulation.

CONTEXT: \\
Agent: \{agent\_name\} (running for mayor) \\
Current situation: "\{context\_summary\}"

TASK: Determine which identity traits are most relevant for this situation.

AVAILABLE RELATIONSHIP TYPES: \\
- profession, years\_experience, is\_politically \\
- prefers\_tech\_adoption\_style, prefers\_planning\_approach \\
- values, believes, has\_experience\_in, led\_project

REASONING GUIDELINES: \\
1. If context mentions "environment/pollution" $\rightarrow$ prioritize: values, believes (environmental) \\
2. If context mentions "technology/innovation" $\rightarrow$ prioritize: prefers\_tech\_adoption\_style, believes (tech) \\
3. If context mentions "community/tradition" $\rightarrow$ prioritize: values (tradition), has\_experience\_in \\
4. If context mentions "economy/development" $\rightarrow$ prioritize: prefers\_planning\_approach, led\_project \\
5. If context mentions "campaign/election" $\rightarrow$ prioritize: is\_politically, values \\
6. Always include: profession, is\_politically (for baseline context)

OUTPUT FORMAT: \\
Return a JSON object with priority-ordered relationship types and optional keywords: \\
\{\\
    "high\_priority": ["relationship\_type1", "relationship\_type2"],\\
    "medium\_priority": ["relationship\_type3"],\\
    "keywords": ["keyword1", "keyword2"]\\
\}

EXAMPLE: \\
For context about pollution: \\
\{"high\_priority": ["values", "believes"], "medium\_priority": ["has\_experience\_in"], "keywords": ["environment", "sustainability"]\}

"""
}
\end{quote}
\end{scriptsize}

\subsection{Agent Bob's Chronicle (Textual Representation)}

\begin{scriptsize}
\begin{quote}
"""
{\ttfamily

Bob is a progressive urban planner with 15 years of experience.

Bob values rapid modernization and technological advancement over environmental sustainability.

Bob began his career developing smart infrastructure systems.
Previously, Bob focused on integrating sensors, data platforms, and automation into city planning.

Bob supports large-scale innovation that increases efficiency and economic performance.

Bob prefers experimental, forward-looking approaches over traditional planning models.

Politically, Bob supports modernist policies.

Bob promotes fast, technology-driven development that favors progress over preservation.

Bob believes legacy infrastructure should be replaced with automated, high-performance systems.

Bob supports policies that encourage innovation, public-private tech partnerships, and global economic competitiveness.

Bob opposes frameworks that delay growth or restrict adoption of new technologies.

Over the years, Bob has led projects that deployed IoT-based infrastructure, designed autonomous transit systems, and implemented real-time data platforms for urban governance.

Bob believes cities should embrace rapid innovation, optimize resource distribution through smart systems, and evolve through continuous experimentation and scalable design.

Bob supports replacing outdated systems with new technology that enables adaptive, high-efficiency urban environments.}

"""

\end{quote}
\end{scriptsize}

\subsection{Agent Alice's Chronicle (Textual Representation)}

\begin{scriptsize}
\begin{quote}
"""
{\ttfamily

Alice is a conservative urban planner with 20 years of experience.

Alice values cultural continuity and historical preservation over technological advancement. 

Alice began her career working in heritage districts.

Previously, Alice focused on protecting historical buildings and adapting infrastructure to modern standards.

Alice supports incremental improvements that are based on local community needs. 

Alice prefers time-tested planning methods over experimental approaches.

Politically, Alice supports preservationist policies.

Alice promotes slow, sustainable development that maintains long-term environmental and economic health.

Alice believes technology should be tested, introduced gradually, and integrated with existing systems.

Alice opposes large-scale modernization efforts that may disrupt cultural identity or the urban landscape.

Alice promotes policies that protect the environment and support economic resilience through traditional infrastructure.

Over the years, Alice has led projects that restored historic architecture, created low-rise zoning plans, and introduced community sustainability programs like recycling and public parks.

Alice believes cities should protect their historical heritage, prioritize ecological sustainability, and build resilience through deliberate planning and community-specific approaches.

Alice supports limited use of technology that enhances existing systems without replacing them.}

"""

\end{quote}
\end{scriptsize}

\end{document}